# Exploiting gradients and Hessians in Bayesian optimization and Bayesian quadrature

**Anqi Wu**  ANQIW@PRINCETON.EDU
*Princeton Neuroscience Institute*
*Princeton University*
*Princeton, NJ 08544, USA*

**Mikio C. Aoi**  MAOI@PRINCETON.EDU
*Princeton Neuroscience Institute*
*Princeton University*
*Princeton, NJ 08544, USA*

**Jonathan W. Pillow**  PILLOW@PRINCETON.EDU
*Princeton Neuroscience Institute*
*Princeton University*
*Princeton, NJ 08544, USA*

## Abstract

An exciting branch of machine learning research focuses on methods for learning, optimizing, and integrating unknown functions that are difficult or costly to evaluate. A popular Bayesian approach to this problem uses a Gaussian process (GP) to construct a posterior distribution over the function of interest given a set of observed measurements, and selects new points to evaluate using the statistics of this posterior. Here we extend these methods to exploit derivative information from the unknown function. We describe methods for Bayesian optimization (BO) and Bayesian quadrature (BQ) in settings where first and second derivatives may be evaluated along with the function itself. We perform sampling-based inference in order to incorporate uncertainty over hyperparameters, and show that both hyperparameters and function uncertainty decrease much more rapidly when using derivative information. Moreover, we introduce techniques for overcoming ill-conditioning issues that have plagued earlier methods for gradient-enhanced Gaussian processes and Kriging. We illustrate the efficacy of these methods using applications to real and simulated Bayesian optimization and quadrature problems, and show that exploiting derivatives can provide substantial gains over standard methods.

**Keywords:** Gaussian process, Bayesian optimization, Bayesian quadrature, gradients, Hessians

## 1. Introduction

An important family of machine learning problems involves adaptively selecting measurement locations in order to infer properties of an unknown function of interest. Common problems include *optimization*, which involves searching for a function's maximum or minimum, and *quadrature*, which involves computing an integral over the unknown function. The Bayesian approach to these problems is to construct a posterior distribution over the function, updated after every observation, which captures all information about any desired statistic (e.g., the location of the optimum, or the value of an integral), and can be used to select new points at which to evaluate the function. Gaussian processes (GPs) provide a convenient and flexible prior distribution over functions, and have therefore formed a central component of Bayesian methods for function learning, optimization, and quadrature (Rasmussen and Williams, 2006; Osborne et al., 2009; Snoek et al., 2012; Wang et al., 2013; Osborne et al., 2012; Gunter et al., 2014).

In the case of function optimization, it is worth noting that there is a dichotomy between classical numerical methods designed to find local optima, and Bayesian optimization methods that focus on global optima. Classical methods like *steepest descent* and *Newton's method* typically rely on gradient (1st derivative) and Hessian (2nd derivative) information to ascend or descend a function locally. Bayesian methods, on the other hand, typically ignore gradient information and rely solely on function evaluations to find an optimum. In cases where gradient information is available, however, this divide is unnatural. GPs provide well-defined probability distributions over the derivatives of a function (for suitable choices of covariance function), making it possible to use observed derivatives as well as function values to update the posterior. As we will show, incorporating such information can provide major advantages for GP-based global optimization and quadrature.

Although many of the functions considered in the Bayesian optimization and quadrature literature are not amenable to this approach because their derivatives cannot be easily evaluated, a variety of objective functions exist for which gradients and Hessians can be calculated relatively cheaply. For example, loss functions involving regularization parameters (Domke, 2012; Larsen et al., 1998; Maclaurin et al., 2015) and SVM kernel parameters (Chapelle et al., 2002) have been shown to allow for tractable calculations of gradients.

In this paper, we focus on closed-form marginal likelihood functions found in GP regression and related models, which are non-convex but continuous with respect to model hyperparameters. For these functions, the terms required for the computing derivatives are byproducts of function evaluation, making it relatively inexpensive to return gradients and Hessians along with the function value (Rasmussen and Williams, 2006; Hensman and Lawrence, 2014). Marginal likelihoods are nonetheless expensive to compute for larger datasets, which (along with non-convexity) motivates the use of Bayesian optimization and quadrature methods.

Previous studies in both the Kriging (Lockwood and Anitescu, 2012; Dalbey, 2013; Banerjee and Gelfand, 2006) and GP literatures (Osborne et al., 2009; Solak et al., 2003; Riihimäki and Vehtari, 2010; Wu et al., 2017) have proposed methods for incorporating derivative information into GPs, but these methods have often been hindered by practical limitations, in particular ill-conditioning of the posterior covariance that often arises when derivative information, especially Hessian, is included. To overcome this problem, we describe two strategies: *input rescaling* and *spectral representation* of the Gaussian process. We show that this spectral representation can be made arbitrarily accurate so that our solution is as close to the exact GP posterior as desired. We demonstrate the superior optimization and quadrature performance of our methods using both deterministic unimodal and multimodal functions, and show a practical application to marginal likelihood optimization. We show that the resulting methods for BO and BQ exhibit faster convergence and smaller uncertainty than standard methods.

The paper is organized as follows. In Section 2, we review Gaussian processes. In Section 3, we describe Bayesian updating of GPs using derivative information. In Section 4, we will introduce input rescaling and spectral representation of GPs with derivatives. In Section 5, we show applications of our methods to realistic BO and BQ problems, and finally, in Section 6, we summarize and discuss the significance of our paper in context of the prior literature.

## 2. Gaussian process priors over functions

Gaussian processes provide a flexible and computationally tractable prior distribution over functions. A GP is parametrized by a mean function $m(\mathbf{x})$ and a covariance function $k(\mathbf{x}, \mathbf{x}')$. A popular choice for covariance function is the *squared exponential* (SE) kernel, $k(\mathbf{x}, \mathbf{x}') = \rho \exp\left(-\frac{||\mathbf{x}-\mathbf{x}'||_2^2}{2\delta^2}\right)$, where $\rho$ controls the marginal variance of function values and $\delta$ is the length scale, which determines the fall-off in correlation with distance $||\mathbf{x} - \mathbf{x}'||$. Functions sampled from a GP with SE covariance are infinitely differentiable, and their smoothness increases with $\delta$.

A GP induces a multivariate Gaussian distribution over any finite collections of function values. If we have a function $f \sim \mathcal{GP}(m, K)$, then for any set of input $N$ locations $\mathbf{x}_{1:N} = \{\mathbf{x}_n \in \mathcal{X}\}_{n=1}^N$ in input domain $\mathcal{X}$, the vector of function values $\mathbf{f}_{1:N} = \{f(\mathbf{x}_n) \in \mathbb{R}\}_{n=1}^N$ has a multivariate Gaussian distribution, $\mathbf{f}_{1:N} \sim \mathcal{N}(\mathbf{m}(\mathbf{x}_{1:N}), \mathbf{K}(\mathbf{x}_{1:N}, \mathbf{x}_{1:N}))$, where $\mathbf{K}$ is the covariance matrix with $i,j$'th element equal to $k(\mathbf{x}_i, \mathbf{x}_j)$. For simplicity, we will assume that the mean $m(\cdot) = 0$, although it is straightforward to incorporate a non-zero mean function if desired.

Given a set of observed function values $\mathcal{D} = \{\mathbf{x}_{1:N}, \mathbf{f}_{1:N}\}$ for a function sampled from a GP, the rules for conditionalization of multivariate Gaussian densities allow us to derive the posterior distribution, which also takes the form of a GP. Let $\mathbf{x}^*$ denote an arbitrary collection of locations and $\mathbf{f}^* = f(\mathbf{x}^*)$ denote the corresponding function values. Then we have:

$$\begin{bmatrix} \mathbf{f}_{1:N} \\ \mathbf{f}^* \end{bmatrix} \sim \mathcal{N}\left(\mathbf{0}, \begin{bmatrix} \mathbf{K} & \mathbf{k}^* \\ \mathbf{k}^{*\top} & k(\mathbf{x}^*, \mathbf{x}^*) \end{bmatrix}\right) \qquad (1)$$

where $\mathbf{k}^* = [k(\mathbf{x}^*, \mathbf{x}_1), k(\mathbf{x}^*, \mathbf{x}_2), \cdots, k(\mathbf{x}^*, \mathbf{x}_N)]^\top$. Thus we have the following posterior distribution over $\mathbf{f}^*$:

$$\mathbf{f}^* | \mathbf{x}_{1:N}, \mathbf{f}_{1:N}, \mathbf{x}^* \sim \mathcal{N}(\mu(\mathbf{x}^*), \sigma^2(\mathbf{x}^*)) \qquad (2)$$

where

$$\begin{aligned} \mu(\mathbf{x}^*) &= \mathbf{k}^{*\top} \mathbf{K}^{-1} \mathbf{f}_{1:N} \\ \sigma^2(\mathbf{x}^*) &= k(\mathbf{x}^*, \mathbf{x}^*) - \mathbf{k}^{*\top} \mathbf{K}^{-1} \mathbf{k}^* \end{aligned} \qquad (3)$$

$\mu(\mathbf{x}^*)$ and $\sigma^2(\mathbf{x}^*)$ are posterior mean and covariance.

## 3. Gaussian processes with derivatives

Gaussian processes with sufficiently smooth covariance functions (e.g., squared exponential or Matérn kernel of order $\nu > 2$) induce a well-defined distribution over functions as well as their first and second derivatives. Here we derive the joint distribution over a function and its derivatives under a GP. This allows for Bayesian updating given observations of a function and its gradient and Hessian.

We consider a $d$-dimensional function sampled from a GP, $f(\cdot) \sim \mathcal{GP}(0, K)$. Let us denote the first order partial derivative operator as $\nabla(\cdot) = \begin{bmatrix} \frac{\partial}{\partial x_1} & \frac{\partial}{\partial x_2} & \cdots & \frac{\partial}{\partial x_d} \end{bmatrix}^\top$. Then the joint process $[f(\mathbf{x}), \nabla f(\mathbf{x})]$ has a GP distribution (see, e.g., Solak et al. (2003); Riihimäki and Vehtari (2010); Banerjee et al. (2003)) with a covariance function given by four blocks:

$$
\begin{aligned}
k_{[f,f]}(\mathbf{x}, \mathbf{x}') &= \operatorname{cov}(f(\mathbf{x}), f(\mathbf{x}')) = k(\mathbf{x}, \mathbf{x}') \\
k_{[f,\nabla f]}(\mathbf{x}, \mathbf{x}') &= \operatorname{cov}(f(\mathbf{x}), \nabla f(\mathbf{x}')) = \nabla_{\mathbf{x}'} k(\mathbf{x}, \mathbf{x}') \\
k_{[\nabla f, f]}(\mathbf{x}, \mathbf{x}') &= \operatorname{cov}(\nabla f(\mathbf{x}), f(\mathbf{x}')) = \nabla_{\mathbf{x}} k(\mathbf{x}, \mathbf{x}') \\
k_{[\nabla f, \nabla f]}(\mathbf{x}, \mathbf{x}') &= \operatorname{cov}(\nabla f(\mathbf{x}), \nabla f(\mathbf{x}')) = \nabla_{\mathbf{x}} \nabla_{\mathbf{x}'} k(\mathbf{x}, \mathbf{x}'),
\end{aligned}
\quad (4)
$$

where the first block is simply the covariance function for the original GP. We can write this joint process more compactly as

$$
\begin{bmatrix} f \\ \nabla f \end{bmatrix} \sim \mathcal{GP}\left( \mathbf{0}, \begin{bmatrix} \mathbf{k} & \mathbf{k}_{[f,\nabla f]} \\ \mathbf{k}_{[\nabla f, f]} & \mathbf{k}_{[\nabla f, \nabla f]} \end{bmatrix} \right). \quad (5)
$$

We can now apply the same conditionalization formulas to derive the posterior over function values $\mathbf{f}^*$ given a set of observations of function values and gradients. Let $\mathbf{K}_{[\mathbf{f},\nabla \mathbf{f}]}$ denote the joint kernel matrix for a set of observations of function values and gradients (from (5)). The joint distribution for $[\mathbf{f}_{1:N}, \nabla \mathbf{f}_{1:N}, \mathbf{f}^*]$ is

$$
\begin{bmatrix} \mathbf{f}_{1:N} \\ \nabla \mathbf{f}_{1:N} \\ \mathbf{f}^* \end{bmatrix} \sim \mathcal{N}\left( \mathbf{0}, \begin{bmatrix} \mathbf{K}_{[\mathbf{f},\nabla \mathbf{f}]} & \bar{\mathbf{k}}^* \\ \bar{\mathbf{k}}^{*\top} & k(\mathbf{x}^*, \mathbf{x}^*) \end{bmatrix} \right), \quad (6)
$$

where $\bar{\mathbf{k}}^* = [\mathbf{k}^{*\top}, \mathbf{k}_{[\mathbf{f}^*,\nabla \mathbf{f}]}]^\top$, and the posterior over $\mathbf{f}^*$ is:

$$
\mathbf{f}^* | \mathbf{x}_{1:N}, [\mathbf{f}, \nabla \mathbf{f}]_{1:N}, \mathbf{x}^* \sim \mathcal{N}(\bar{\mu}(\mathbf{x}^*), \bar{\sigma}^2(\mathbf{x}^*)) \quad (7)
$$

where

$$
\begin{aligned}
\bar{\mu}(\mathbf{x}^*) &= \bar{\mathbf{k}}^{*\top} \mathbf{K}_{[\mathbf{f},\nabla \mathbf{f}]}^{-1} [\mathbf{f}, \nabla \mathbf{f}]_{1:N}^\top \\
\bar{\sigma}^2(\mathbf{x}^*) &= k(\mathbf{x}^*, \mathbf{x}^*) - \bar{\mathbf{k}}^{*\top} \mathbf{K}_{[\mathbf{f},\nabla \mathbf{f}]}^{-1} \bar{\mathbf{k}}^*
\end{aligned}
\quad (8)
$$

We use a similar derivation to incorporate Hessian (2nd derivative) information into a GP along with the function and gradient information. For GPs with observations of the Hessian, we write the second order partial derivative operator (which contains $d^2$ elements) as

$$
\nabla \otimes \nabla^\top(\cdot) = \begin{bmatrix} \frac{\partial^2}{\partial x_1 \partial x_1} & \cdots & \frac{\partial^2}{\partial x_1 \partial x_d} \\ \vdots & \ddots & \vdots \\ \frac{\partial^2}{\partial x_d \partial x_1} & \cdots & \frac{\partial^2}{\partial x_d \partial x_d} \end{bmatrix} \quad (9)
$$

which can be rearranged into a column vector of $d(d+1)/2$ elements (keeping only unique terms):

$$\nabla^2(\cdot) = \begin{bmatrix} \frac{\partial^2}{\partial x_1 \partial x_1} & \frac{\partial^2}{\partial x_2 \partial x_1} & \cdots & \frac{\partial^2}{\partial x_d \partial x_1} & \cdots & \frac{\partial^2}{\partial x_{d-1} \partial x_d} & \frac{\partial^2}{\partial x_d \partial x_d} \end{bmatrix}^\top$$

The joint process $[f(\mathbf{x}), \nabla f(\mathbf{x}), \nabla^2 f(\mathbf{x})]$ has a valid stationary cross-covariance matrix function, apart from (4),

$$\begin{aligned}
k_{[f,\nabla^2 f]}(\mathbf{x}, \mathbf{x}') &= \mathrm{cov}(f(\mathbf{x}), \nabla^2 f(\mathbf{x}')) = \nabla^2_{\mathbf{x}'} k(\mathbf{x}, \mathbf{x}') \\
k_{[\nabla^2 f, f]}(\mathbf{x}, \mathbf{x}') &= \mathrm{cov}(\nabla^2 f(\mathbf{x}), f(\mathbf{x}')) = \nabla^2_{\mathbf{x}} k(\mathbf{x}, \mathbf{x}') \\
k_{[\nabla f, \nabla^2 f]}(\mathbf{x}, \mathbf{x}') &= \mathrm{cov}(\nabla f(\mathbf{x}), \nabla^2 f(\mathbf{x}')) = \nabla_{\mathbf{x}} \nabla^2_{\mathbf{x}'} k(\mathbf{x}, \mathbf{x}') \\
k_{[\nabla^2 f, \nabla f]}(\mathbf{x}, \mathbf{x}') &= \mathrm{cov}(\nabla^2 f(\mathbf{x}), \nabla f(\mathbf{x}')) = \nabla^2_{\mathbf{x}} \nabla_{\mathbf{x}'} k(\mathbf{x}, \mathbf{x}') \\
k_{[\nabla^2 f, \nabla^2 f]}(\mathbf{x}, \mathbf{x}') &= \mathrm{cov}(\nabla^2 f(\mathbf{x}), \nabla^2 f(\mathbf{x}')) = \nabla^2_{\mathbf{x}} \nabla^2_{\mathbf{x}'} k(\mathbf{x}, \mathbf{x}')
\end{aligned} \quad (10)$$

Then the the joint Gaussian of $[f, \nabla f, \nabla^2 f(\mathbf{x})]$ is defined as,

$$\begin{bmatrix} f \\ \nabla f \\ \nabla^2 f \end{bmatrix} \sim \mathcal{GP}\left(\mathbf{0}, \mathbf{K}_{[\mathbf{f}, \nabla \mathbf{f}, \nabla^2 \mathbf{f}]}\right) \quad (11)$$

where

$$\mathbf{K}_{[\mathbf{f}, \nabla \mathbf{f}, \nabla^2 \mathbf{f}]} \equiv \begin{bmatrix} \mathbf{k} & \mathbf{k}_{[f, \nabla f]} & \mathbf{k}_{[f, \nabla^2 f]} \\ \mathbf{k}_{[\nabla f, f]} & \mathbf{k}_{[\nabla f, \nabla f]} & \mathbf{k}_{[\nabla f, \nabla^2 f]} \\ \mathbf{k}_{[\nabla^2 f, f]} & \mathbf{k}_{[\nabla^2 f, \nabla f]} & \mathbf{k}_{[\nabla^2 f, \nabla^2 f]} \end{bmatrix} \quad (12)$$

We can now apply the same conditionalization formulas to derive the posterior over function values $\mathbf{f}^*$ given a set of observations of function values, gradients and Hessians. The joint distribution of $[\mathbf{f}_{1:N}, \nabla \mathbf{f}_{1:N}, \nabla^2 \mathbf{f}_{1:N}, \mathbf{f}^*]$ is

$$\begin{bmatrix} \mathbf{f}_{1:N} \\ \nabla \mathbf{f}_{1:N} \\ \nabla^2 \mathbf{f}_{1:N} \\ \mathbf{f}^* \end{bmatrix} \sim \mathcal{N}\left(\mathbf{0}, \begin{bmatrix} \mathbf{K}_{[\mathbf{f}, \nabla \mathbf{f}, \nabla^2 \mathbf{f}]} & \bar{\mathbf{k}}^* \\ (\bar{\mathbf{k}}^*)^\top & k(\mathbf{x}^*, \mathbf{x}^*) \end{bmatrix}\right), \quad (13)$$

where $\bar{\mathbf{k}}^* = [\mathbf{k}^{*\top}, \mathbf{k}_{[\mathbf{f}^*, \nabla f]}, \mathbf{k}_{[\mathbf{f}^*, \nabla^2 f]}]^\top$, and the posterior over $\mathbf{f}^*$ is:

$$\mathbf{f}^* | \mathbf{x}_{1:N}, [\mathbf{f}, \nabla \mathbf{f}, \nabla^2 \mathbf{f}]_{1:N}, \mathbf{x}^* \sim \mathcal{N}(\bar{\mu}(\mathbf{x}^*), \bar{\sigma}^2(\mathbf{x}^*)) \quad (14)$$

where

$$\begin{aligned}
\bar{\mu}(\mathbf{x}^*) &= (\bar{\mathbf{k}}^*)^\top \mathbf{K}_{[\mathbf{f}, \nabla \mathbf{f}, \nabla^2 \mathbf{f}]}^{-1} [\mathbf{f}, \nabla \mathbf{f}, \nabla^2 \mathbf{f}]_{1:N}^\top \\
\bar{\sigma}^2(\mathbf{x}^*) &= k(\mathbf{x}^*, \mathbf{x}^*) - (\bar{\mathbf{k}}^*)^\top \mathbf{K}_{[\mathbf{f}, \nabla \mathbf{f}, \nabla^2 \mathbf{f}]}^{-1} \bar{\mathbf{k}}^*
\end{aligned} \quad (15)$$

The resulting joint kernel matrix is partitioned into 9 blocks corresponding to the covariances and cross-covariances over function values, gradients and Hessians.

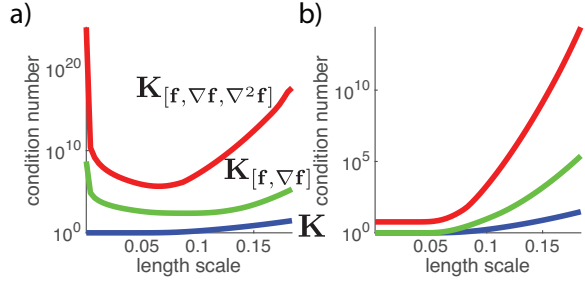

Figure 1: a) Condition number of the SE covariance matrix $\mathbf{K}$ for a 1D function after 100 observations spaced 0.2 unit apart, as a function of length scale $\delta$. b) Condition number after applying the rescaling trick (note the different scales of the y-axes).

## 4. Ill-conditioning

Both Bayesian optimization and quadrature exploit smooth covariance kernels, e.g., squared exponential or Matérn kernel of order $\nu > 2$, etc. Most of these kernels rely on $\frac{||\mathbf{x}-\mathbf{x}'||}{\delta}$, thus are stationary which is translation invariant. Larger values of $\delta$ result in a smoother function, while smaller values encourage fluctuations. For this paper, we mainly focus on the squared exponential kernel, but the derivative-enhanced method can be applied to any kernel for Bayesian learning. More analyses and applications to the Matérn 5/2 kernel will be found in the appendix.

There is undoubted value in incorporating derivative information when it is available, but the SE kernel K becomes ill-conditioned much more rapidly when using these measurements (see Fig. 1 a)). When $\delta$ is small, blocks of the kernel for Hessians will explode due to the multiplication of $1/\delta$. When $\delta$ is large (i.e., smooth functions), Hessian-based methods lead to much faster ill-conditioning as the matrix gets more singular. A spectral (Fourier) representation of the GP alleviates this problem. We will further analyze the two extreme cases for the ill-conditioned kernels and provide two solutions, a rescaling method and a spectral representation, for each scenario.

### 4.1 Input rescaling

From (4), we observe that the covariance between two derivatives of $\mathbf{f}$ is just the derivatives of the covariance function $k(\mathbf{x}, \mathbf{x}')$ w.r.t. $\mathbf{x}$ and $\mathbf{x}'$. However, it is obvious that the higher order derivatives of $k(\mathbf{x}, \mathbf{x}')$ would incur a large multiplication of $\frac{1}{\delta}$, up to $\frac{1}{\delta^8}$ for Hessian, which induces numerical explosion when $\delta < 1$. This a general issue for most of the kernels containing the $\exp(\frac{||\mathbf{x}-\mathbf{x}'||}{\delta})$ term. This results in the rapidly increasing condition number on the left-side shown in Fig. 1.

To cope with this situation, we propose a rescaling method which simply absorbs $\delta$ into $\mathbf{x}$ before taking derivatives, i.e. $\widetilde{\mathbf{x}} = \mathbf{x}/\delta$. This is based on the intuition that transforming the distance $||\mathbf{x}-\mathbf{x}'||$ into unit scale with $\delta = 1$ will not hurt the original covariance function $\mathbf{k}$, while weakening the significance of gradients and Hessians for the posterior estimation.

In practice, there is also the possibility of measurement noise $\epsilon$, which we will assume is Gaussian, $\epsilon \sim \mathcal{N}(0, \sigma_n^2)$, and sample-dependent. If the noise is additive, we can easily add the noise distribution to the Gaussian distribution and define $\mathbf{y}_i = f(\mathbf{x}_i) + \epsilon_i$, which implies that we can add the noise to the original kernel $\mathbf{K}$, $\mathbf{K}_{noise} = \mathbf{K} + \sigma_n^2 I$. Then correspondingly, the posterior mean and variance are altered as follows,

$$\bar{\mu}(\mathbf{x}^*) = \bar{\mathbf{k}}^{*\top}(\mathbf{K}_{[\mathbf{f},\nabla\mathbf{f}]} + \sigma_n^2 I)^{-1}[\mathbf{f},\nabla\mathbf{f}]_{1:N}^{\top}, \; \bar{\sigma}^2(\mathbf{x}^*) = k(\mathbf{x}^*,\mathbf{x}^*) - \bar{\mathbf{k}}^{*\top}(\mathbf{K}_{[\mathbf{f},\nabla\mathbf{f}]} + \sigma_n^2 I)^{-1}_{[\mathbf{f},\nabla\mathbf{f}]}\bar{\mathbf{k}}^* \quad (16)$$

With the rescaling method, we need to rescale $\bar{\mathbf{k}}^*$, $\mathbf{K}_{[\mathbf{f},\nabla\mathbf{f}]}$ and $[\mathbf{f},\nabla\mathbf{f}]_{1:N}^\top$ in order to achieve a reasonable posterior. Since we rescale the distance among $\mathbf{x}$, keeping the entire function space isometric, we need to rescale their function values $\mathbf{f}$ as well. More intuitively, we notice that the SE kernel will separate out a $\frac{1}{\delta}$ every time it's differentiated, thus the order of $\frac{1}{\delta}$ in front of the exponential should be proportional to the order of its derivatives. It's easy to prove that rescaling every term (except for the noise term) in (16) for the SE kernel is equivalent to retaining everything but rescaling the noise according to the order of the derivatives in each block of the kernel. This is consistent with a general assumption for noisy GPs that large observation values should be assigned larger environmental noise. Thus the rescaling technique mitigates the ill-condition (i.e. small $\delta$) since we suppress the very large singular values of the kernel (Fig. 1 b)).

### 4.2 Spectral domain representation

When $\delta$ gets larger in Fig. 1, the SE kernel becomes ill-conditioned due to the strong singularity. From a spectral point of view, the singular values correspond to the effective frequencies in the frequency domain. Instead of cutting off small singular values with a hard threshold, which usually leads to a non-smooth function curve as a function of $\delta$, we provide a spectral representation of the covariance kernel which effectively transforms the ill-conditioned dual (original) form to a low-dimensional primal (spectral) form. In addition, a new posterior update rule is derived without the need to explicitly partition the kernel matrix into separate blocks for function values, gradients, Hessians and cross terms. This only works for stationary kernels. The novelty of our spectral kernel resides in three aspects: a controllable approximation accuracy with arbitrary precision (e.g., to floating point accuracy, and beyond, if desired); a neat and convenient derivation of the kernel with derivatives; and, a new framework for spectral Bayesian optimization and quadrature in the frequency domain, which resolves the ill-condition (i.e. large $\delta$) of the kernel.

For a stationary Gaussian process, the kernel function is entirely specified by its autocorrelation function $k(\boldsymbol{\tau}) = k(\mathbf{x}, \mathbf{x}')$, where $\boldsymbol{\tau} = \mathbf{x} - \mathbf{x}'$ is the separation between any pair of locations. Bochner's theorem (Bochner, 2016) states that any stationary covariance function can be represented as the Fourier transform of a positive finite measure, which corresponds to the power spectral density $s(\boldsymbol{\omega}) = \int e^{-2\pi i \boldsymbol{\omega}^\top \boldsymbol{\tau}} k(\boldsymbol{\tau}) d\boldsymbol{\tau}$. For the SE kernel we have

$$s(\boldsymbol{\omega}) = (2\pi\delta^2)^{p/2} \rho \exp\left(-2\pi^2 \delta^2 \boldsymbol{\omega}^2\right) \tag{17}$$

where $p$ is the dimension of the input space. This entails that the GP has an equivalent representation in the Fourier domain with a diagonal covariance, namely:

$$g \sim \mathcal{N}(\alpha(\boldsymbol{\omega}), \Sigma(\boldsymbol{\omega})) \tag{18}$$

where $g$ is the Fourier transform of $f$, $\alpha(\boldsymbol{\omega})$ is the Fourier transform of the mean function $m(\mathbf{x})$, and $\Sigma = \text{diag}(s(\boldsymbol{\omega}))$ is diagonal, meaning that Fourier components are a priori independent, with prior variance $s(\boldsymbol{\omega})$.

We can obtain an efficient Fourier-domain representation of the GP using the discrete Fourier transform, as discussed in prior work on spectral GPs (Wikle, 2002; Royle and Wikle, 2005; Paciorek, 2007b,a). In practice, this requires picking a lowest frequency $\omega_0$, which will determine a periodic boundary condition for the function, and a highest frequency $\omega_c$ for some integer $c$, resulting in a set of $2c+1$ frequencies: $\boldsymbol{\omega} = \{0, \pm\omega_0, \ldots \pm c\omega_0\}$.

There are two sources of approximation error to be considered. One is related to the periodic boundary condition induced by the discrete Fourier representation. This can be reduced to $\exp(-t^2/2)$ by setting the lowest frequency $\omega_0$ to $\frac{1}{T+t\delta}$, where $T$ is the desired support of $f$, $t$ is a parameter controlling error size and $\delta$ is the length scale of the SE kernel. The other source of error is related to the cutoff of high frequencies, which is the critical step to resolve the ill-conditioning issue. This can also be controlled by setting the highest frequency integer $c$ based on its prior variance $s(c\omega_0)$. We can set the condition number of the prior covariance matrix to $10^{14}$ if we set $c = \sqrt{\frac{14\log 10}{2\pi\delta^2\omega_0^2}}$, meaning we keep all Fourier modes that are at least $10^{-14}$ times as large as the largest mode under the prior. Note that we can control the accuracy of this representation to arbitrary precision (e.g., to floating point accuracy, and beyond, if desired), so we emphasize that this approach can be considered to support exact and not merely approximate GP inference (*cf.* Lázaro-Gredilla et al. (2010)).

The mapping between $f(\mathbf{x})$ and its Fourier transform $g(\boldsymbol{\omega})$ is therefore given by

$$f(\mathbf{x}) = \sum_j e^{2\pi i \boldsymbol{\omega}_j^\top \mathbf{x}} g(\boldsymbol{\omega}_j) = B^\top g(\boldsymbol{\omega}) \tag{19}$$

where $B$ is a column vector with entries $e^{2\pi i \boldsymbol{\omega}_j^\top \mathbf{x}}$ on the $j^{th}$ position. Given a set of observation pairs $\mathcal{D} = \{\mathbf{x}_{1:N}, \mathbf{f}_{1:N}\}$, we have $\mathbf{f}_{1:N} = B_{1:N}^\top g(\boldsymbol{\omega})$ where the $i$th column of $B_{1:N}$ represents the discrete Fourier transform base from $g(\boldsymbol{\omega})$ to $f(\mathbf{x}_i)$. We omit the subscript of $B_{1:N}$ for simplicity. One caveat is we generally don't assume uniform samples for Bayesian learning (e.g., optimization and quadrature), thus $B$ is a non-uniform DFT from gridded frequencies $\boldsymbol{\omega}$ to real valued $\mathbf{x}$. ($B$ is therefore not orthogonal).

To derive the GP representation of $g$, we start with the joint Gaussian distribution for $g$ and $\mathbf{f}_{1:N}$,

$$\begin{bmatrix} \mathbf{f}_{1:N} \\ g \end{bmatrix} \sim \mathcal{N}\left( \begin{bmatrix} B^\top \alpha \\ \alpha \end{bmatrix}, \begin{bmatrix} B^\top \Sigma B & B^\top \Sigma \\ \Sigma B & \Sigma \end{bmatrix} \right) \tag{20}$$

where $B^\top \Sigma B = \mathbf{K}$. If we have noise in our observations of $f$, then $f(\mathbf{x}) = B^\top g(\boldsymbol{\omega}) + \epsilon$, where $\epsilon \sim \mathcal{N}(0, \sigma_n^2)$ is the additive noise. One can also easily derive the posterior over $g$ in the noisy setting,

$$g|\mathbf{f}_{1:N} \sim \mathcal{N}(\widetilde{\alpha}(\boldsymbol{\omega}), \widetilde{\Sigma}(\boldsymbol{\omega})) \tag{21}$$

where

$$\begin{aligned} \widetilde{\alpha}(\boldsymbol{\omega}) &= \alpha + \Sigma B (B^\top \Sigma B + \sigma_n^2 I)^{-1}(\mathbf{f}_{1:N} - B^\top \alpha) \\ &= \alpha + \frac{1}{\sigma_n^2}(\Sigma^{-1} + \frac{1}{\sigma_n^2} BB^\top)^{-1} B(\mathbf{f}_{1:N} - B^\top \alpha) \\ \widetilde{\Sigma}(\boldsymbol{\omega}) &= \Sigma - \Sigma B(B^\top \Sigma B + \sigma_n^2 I)^{-1} B^\top \Sigma = (\Sigma^{-1} + \frac{1}{\sigma_n^2} BB^\top)^{-1} \end{aligned} \tag{22}$$

This formulation of the posterior suggests a convenient update when fixing the hyper-parameters: it is not necessary to keep track of all observations but only the most recent one. All of the information about previous observations is stored in the posterior distribution of $g$. Every update based upon a new observation leads to a new rank-1 matrix added to the existing $BB^\top$ matrix. For prediction of $f^*$ at $\mathbf{x}^*$, we just apply the deterministic inverse Fourier transform,

$$\widetilde{\mu}(\mathbf{x}^*) = B^{*\top}\widetilde{\alpha}(\boldsymbol{\omega}), \qquad \widetilde{\sigma}^2(\mathbf{x}^*) = B^{*\top}\widetilde{\Sigma}(\boldsymbol{\omega})B^* \tag{23}$$

where $B^*$ is a column vector of basis functions at $\mathbf{x}^*$ with entries $e^{2\pi i \boldsymbol{\omega}_j^\top \mathbf{x}^*}$.

The corresponding spectral GP with gradient and Hessian observations is also easier to derive than the dual form in the real domain. Based on (4), the covariance between $f$ and $\nabla f$ is a partial derivative of $\mathbf{k}$ only with respect to $\mathbf{x}$. In the spectral framework, $k(\mathbf{x}, \mathbf{x}') = B(\mathbf{x})^\top \Sigma B(\mathbf{x}')$, where only the Fourier basis $B$ is a function of $\mathbf{x}$. Therefore, instead of calculating the derivatives for a specific kernel, we only need to obtain the derivatives of $B$ w.r.t. $\mathbf{x}$ and apply $B$ to *any* stationary kernel. Denote $\nabla B = 2\pi i \boldsymbol{\omega} B$ and $\nabla^2 B = -4\pi^2 \boldsymbol{\omega}^2 B$ to be its first and second order derivatives w.r.t. $\mathbf{x}$, then $\nabla f = \nabla B^\top g(\boldsymbol{\omega})$ and $\nabla^2 f = \nabla^2 B^\top g(\boldsymbol{\omega})$. Thus $p(g, \nabla f)$ and $p(g, \nabla^2 f)$ have similar joint distributions as in (20), only with $B$ replaced by $\nabla B$ and $\nabla^2 B$ respectively. This is true for the posteriors $p(g|\nabla f)$ and $p(g|\nabla^2 f)$ in (21) as well.

In addition, we note that $[f, \nabla f, \nabla^2 f]$ are conditionally independent given $g$, thus there is no need to construct a full joint kernel matrix. The joint posterior of $g$ is simply $p(g|f, \nabla f, \nabla^2 f) = p(g|f) p(g|\nabla f) p(g|\nabla^2 f)$. In our implementation, we simply extend the $B$ matrix into $[B, \nabla B, \nabla^2 B]$ in (22) due to the elegant multiplication property of the Gaussian distribution. It can be easily shown that the marginal distribution of $[f, \nabla f, \nabla^2 f]$ over $g$ has the exact same joint covariance kernel as $\mathbf{K}_{[f, \nabla f, \nabla^2 f]}$. Therefore, the spectral GP can not only simplify derivations but also eliminate the book-keeping demands of maintaining a 9-block GP covariance matrix.

### 4.2.1 COMPUTATIONAL COSTS FOR SPECTRAL GP

One bottleneck of the spectral GP with derivatives is its storage complexity with multidimensional inputs. Suppose the input $\mathbf{x}$ is a multidimensional vector with length $d$, the number of frequencies is $p$ along each dimension, and $N$ is the number of observations. Operating the kernel and basis function in the spectral formulation has a memory complexity of $O(p^{2d})$ for all cases, while the dual formulation requires $O(N^2 d^2)$ with gradient only and and $O(N^2 d^4)$ with both gradient and Hessian. For the implementation, we retain both primal and dual formulations and switch based on specific situations. But fortunately, we mostly resort to the spectral GP when $\delta$ is large and thus $p$ is small.

## 5. Applications

### 5.1 Bayesian optimization

GPs offer a powerful method to perform Bayesian optimization (Mockus, 1994). In this section, we compare our methods (GPBOgrad and GPBOhess) with ordinary Gaussian process Bayesian optimization (GPBO) on a variety of standard test functions. In the GPBO framework, observations of the test function are typically made sequentially where, for each iteration, the test point is selected by the maximum of an acquisition function. However, the acquisition function depends on the hyper-parameters of the kernel $(\rho, \delta)$. We therefore use MCMC to approximate a posterior distribution over the hyper-parameters with a Gamma hyperprior, and select points based on the expectation of the acquisition function over the posterior distribution of hyper-parameters (Snoek et al., 2012). For all examples that we present, we use either the expected improvement (EI) (Mockus et al., 1978) or the upper confidence bound (UCB) (Srinivas et al., 2009) as the acquisition function.

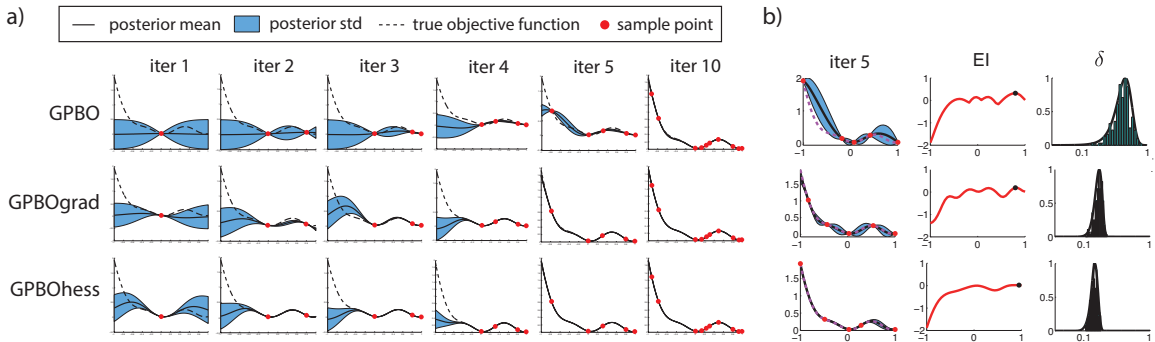

Figure 2: a) Comparison of GPBO, GPBOgrad and GPBOhess with the same set of uniformly random samples and constant GP hyper-parameters. The function is a modified and rescaled Branin-Hoo function with each input range within $[-1, 1]$ and the second input $x_2 = -0.5$. GPBOhess and GPBOgrad shrink around the optimal region after 3 samples; GPBO takes more than 5 samples. For the shrinkage of the entire function, GPhess and GPgrad take 5 evaluations but GP needs 10. b) Comparisons with 5 samples from active sampling selected by the EI acquisition function and the varying length scale $\delta$.

### 5.1.1 UNIMODAL AND MULTIMODAL FUNCTIONS

As an illustrative example, we demonstrate the properties of function estimation and optimization using the 1D Branin-Hoo function. The Branin-Hoo function is a common benchmark for Bayesian optimization techniques (Jones, 2001) that is defined over $x \in \mathcal{R}^2$ where $0 \leq x_1 \leq 15$ and $-5 \leq x_2 \leq 15$ and has three global minima. Here we use a modified form (Forrester et al., 2008) that has two local minima and one global minimum.

First, in order to examine the influence of derivative information on the uncertainty of the function estimate, we present an example where each method is sequentially evaluated by the same set of randomly chosen (not optimized) points with the same values for hyper-parameters. Results in Fig. 2 a) show that for each evaluation of the test function, the more derivative information is available, and the faster the function is learned. For example, note that GPhess has the smallest uncertainty given the same set of samples at each iteration. The Hessian, being a function of the curvature, has information about the location of distant points along the function, which contributes to faster learning and smaller uncertainty about the points on the function near the point that was evaluated.

Another advantage of using observations of the derivatives is in learning the hyper-parameters. For example, consider GPBO of the Branin-Hoo function where EI is averaged over samples from the posterior hyper-parameter distribution. Fig. 2 b) illustrates the 5th iteration of optimization for GPBO, GPBOgrad and GPBOhess, along with the corresponding average the EI and hyper-parameter distribution. We not only find that GPBOhess has already identified the global minimum, but the posterior distribution of $\delta^2$ has smaller variance, which indicates that observing the Hessian provides more information about the hyper-parameters as well. This is advantageous because the faster narrowing of the posterior of the hyper-parameters results in a more precise acquisition function for a given number of observations.

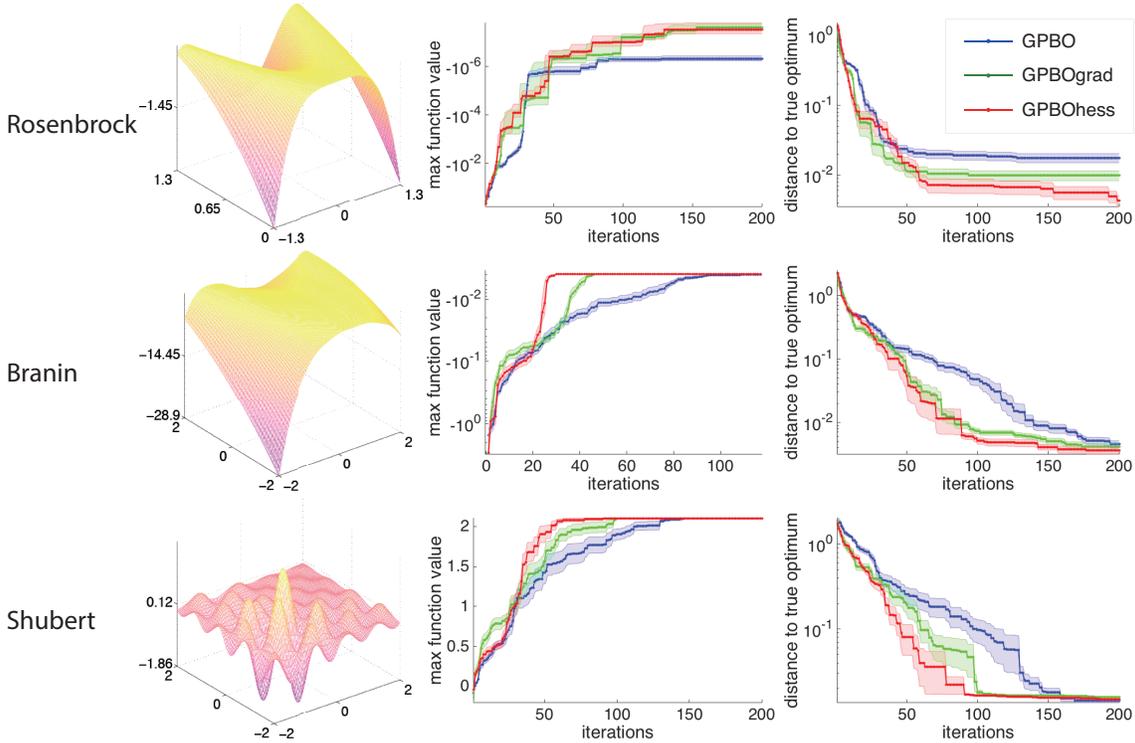

Figure 3: Performance of GP, GPgrad and GPhess for Bayesian optimization with the UCB acquisition function for three functions. With higher order curvature, the method discovers the global optimum location faster and the distance value decreases faster as well.

We also systematically analyzed the performance of GPgrad and GPhess for Bayesian optimization in 2D situations. We applied GPBO for each method to three test functions:

- *Rosenbrock (Banana)*, a unimodal function: $f(\mathbf{x}) = \sum_{i=1}^{d-1}[100(x_{i+1} - x_i^2)^2 + (x_i - 1)^2]$
- *Branin*, a multimodal function: $f(\mathbf{x}) = (x_2 - \frac{5}{4\pi^2}x_1^2 + \frac{5}{\pi}x_1 - 6)^2 + 10(1 - \frac{1}{8\pi})cos(x_1) + 10$
- *Shubert*, a multimodal function with many local minima:
$f(\mathbf{x}) = (\sum_{i=1}^{5} icos((i+1)x_1 + i))(\sum_{i=1}^{5} icos((i+1)x_2 + i))$

The three functions are illustrated in Fig. 3 (left column). We compared the performance of each method over 20 trials by tracking the maximal function value and the minimal distance to the true optimal location. In Fig. 3 (middle and right columns), we can see that GPBOhess converges to the global maximum of all three functions the fastest, while the standard GPBO converges the slowest. In particular, for the Rosenbrock banana, GPBOhess achieves the maximum in nearly half the number of iterations in GPBO. Note that the distance to the global optimum converges particularly fast for GPBOhess compared to GPBO. This suggests that the posterior mean in GPBOhess approximates the function surface in the neighborhood of the global optimum faster and more stably than the other methods. This is critical for Bayesian optimization, where we care less about *what* the optimum is and more about *where* the optimum is.

Notably, standard optimization methods for analytic functions are not extensively elaborated and compared here. This is because we mainly reside in the regime of efficient global optimization methods like Bayesian optimization. However, we do optimize these functions with L-BFGS implemented in *minFunc*[1]. It is obvious that L-BFGS will outperform Bayesian optimization in the first two examples (the estimated maximum function value for Rosenbrock is $5.20e^{-9} \pm 6.46e^{-8}$, and for Branin is $0.40e^{-2} \pm 1.06e^{-10}$), but it will get easily stuck in a local optimum in the Shubert function (the estimated maximum function value = $-0.55 \pm 0.54$). In real applications, a majority of problems get caught up in non-convexity or non-concavity embarrassment, where the global optimization is more needed than local optimization. Methods like optimization algorithms combined with local gradient descent/Newton steps, Newton's method with random restarts or AIS using HMC, could also render a global-like solution, but compared with Bayesian optimization, those are far less practical and efficient. We put our emphasis on the global optimization with less overhead, faster convergence and better guarantee. Admittedly, applying optimization algorithms like Newton's method and its variations could possibly outperform the GPBO on some of the 2D analytic functions. But in consideration of practical evaluation consumption, GPBO regains the right to speak. Therefore, we aim to show a better enhancement of GPBO within the regime of Bayesian optimization, instead of comparing horizontally across all the state-of-art optimization algorithms.

5.1.2 EVIDENCE OPTIMIZATION WITH A SMOOTH PRIOR

Evidence optimization is a popular Bayesian maximum-likelihood procedure for estimating the posterior distribution from data. It's a very effective hyper-parameter learning method for Bayesian regression. Given different properties and structures of the data, people impose different priors controlled by some hyper-parameters. Each evaluation of the objective function in Bayesian regression could be costly when the data resides in a large-scale and high-dimensional space. Fortunately, each of these priors returns closed form derivatives w.r.t. their hyper-parameters, allowing our GPBO methods to give an efficient boost to these problems. In this paper, we applied all GPBO methods to evidence optimization with an ASD prior. ASD models the regression coefficients with a zero-mean Gaussian prior controlled by a pair of hyper-parameters governing the smoothness and the marginal variance of the coefficients. It assigns a non-diagonal prior covariance to the weight vector, given by a Gaussian kernel $C_{ij} = \exp(-r - \Delta_{ij}/2l^2)$ where $\Delta_{ij}$ is the squared distance between two filter coefficients in pixel space. We also estimate the noise variance $\sigma^2$ from the data likelihood. Finally, we used BO to maximize the evidence function for the hyper-parameters $\boldsymbol{\theta} = \{r, l, \sigma^2\}$.

We evaluate the performance using the maximum of the marginal log-likelihood after each iteration. Fig. 4 a-c) show the values of the maximum likelihood at each iteration for the ASD model for three different sizes of the data. But the function values are plotted against the running time in second, instead of iterations. The running time for GPBO is only spent by the objective function evaluation, and the running time for GPBOgrad includes the gradient evaluation. These results indicate that computing the derivatives might lead to some overheads, but as motivated in the beginning, these overheads won't substantially break down the practicability of our derivative-enhanced GPBO methods. Moreover, we also run the L-BFGS for comparison which can possibly find the optimum in short time, but the optimization is highly unstable due to the random initialization. The mean values of the estimated maximum likelihoods and their standard deviations for L-BFGS for the three datasets are $-1.58e^4 \pm 1.66e^3$, $-3.42e^4 \pm 4.10e^3$ and $-7.67e^3 \pm 8.70e^2$, which are certainly way off the true optimal values due to some singular estimations. For fairness, we also compare to the Rasmussen's minimize function[2] which is widely used in the experiments of marginal likelihood optimization, the same local-optimum issue occurs with mean estimations at $-6.16e^4 \pm 8.11e^2$, $-1.35e^5 \pm 1.83e^4$ and $-1.91e^4 \pm 3.55e^3$ respectively, which are bad too.

---

1. http://www.cs.ubc.ca/~schmidtm/Software/minFunc.html

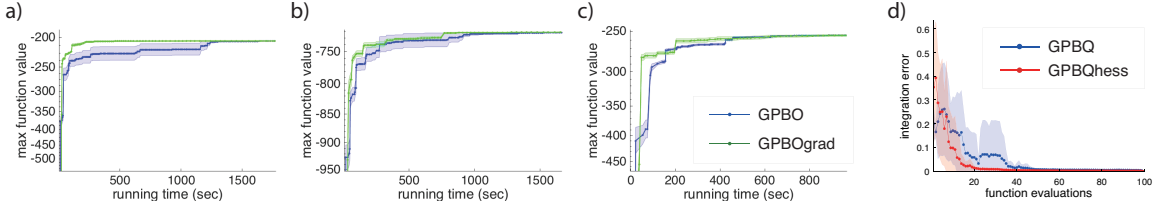

Figure 4: $n$=number of samples, $d$=dimension: a) $n = 8000, d = 50000$; b) $n = 30000, d = 50000$; c) $n = 10000, d = 75000$. d) Bayesian quadrature: the goal is to compute the integral over the marginal likelihood.

### 5.2 Bayesian quadrature

Similarly to BO, Bayesian quadrature (BQ) is used to numerically estimate an integral when function evaluations are prohibitively expensive or intractable. For example, when estimating the marginal likelihood over hyper parameters, the integral is usually denoted as $Z = \int f(\mathbf{x})p(\mathbf{x})d\mathbf{x}$, where the input domain is $\mathcal{X} \to \mathbb{R}^d$, $f(\mathbf{x})$ is the likelihood function and $p(\mathbf{x})$ is the prior density (typically assumed to be a Gaussian). Generally, BQ specifies a prior distribution over $f(\mathbf{x})$ in the form of a Gaussian process (GP). As in (22), given a sample set $\mathcal{D}$, the posterior mean and covariance can be expressed as

$$\widetilde{\mu}(\mathbf{x}) = B(\mathbf{x})^\top \widetilde{\alpha}(\boldsymbol{\omega}), \quad \widetilde{\sigma}^2(\mathbf{x}, \mathbf{x}') = B(\mathbf{x})^\top \widetilde{\Sigma}(\boldsymbol{\omega}) B(\mathbf{x}') \qquad (24)$$

where $B(\mathbf{x})$ is a Fourier basis vector at $\mathbf{x}$. Because the integration of the likelihood is a linear projection onto a Gaussian prior, the posterior distribution of the integral is also Gaussian, yielding a posterior mean and variance of the integral that can be expressed analytically. Denote $Z_B := \int B(\mathbf{x})p(\mathbf{x})d\mathbf{x}$, then the moments of the integral $Z$ are given by,

*posterior mean of integral:*

$$\mathbb{E}(Z|\mathcal{D}) = \int \widetilde{\mu}(\mathbf{x})p(\mathbf{x})d\mathbf{x} = \left(\int B(\mathbf{x})p(\mathbf{x})d\mathbf{x}\right)^\top \widetilde{\alpha}(\boldsymbol{\omega}) = Z_B^\top \widetilde{\alpha}(\boldsymbol{\omega}) \qquad (25)$$

---

2. http://learning.eng.cam.ac.uk/carl/code/minimize/minimize.ml

*posterior variance of integral:*

$$\begin{aligned}\mathbb{V}(Z|\mathcal{D}) &= \int\int \widetilde{\sigma}^2(\mathbf{x},\mathbf{x}')p(\mathbf{x})p(\mathbf{x}')\mathrm{d}\mathbf{x}\mathrm{d}\mathbf{x}' \\ &= \Big(\int B(\mathbf{x})p(\mathbf{x})\mathrm{d}\mathbf{x}\Big)^\top \widetilde{\Sigma}(\boldsymbol{\omega})\Big(\int B(\mathbf{x}')p(\mathbf{x}')\mathrm{d}\mathbf{x}'\Big) \\ &= Z_B^\top \widetilde{\Sigma}(\boldsymbol{\omega})Z_B\end{aligned} \quad (26)$$

Given that the mean and variance above have simple analytical expressions, the GP framework can assist in approximating the marginal likelihood when the likelihood is expensive to evaluate. Of course, this is provided that a principled active sampling strategy is employed to both minimize the uncertainty of the integral and control the number of function evaluations required to achieve a given tolerance. Suppose we denote the next best location as $\mathbf{x}^*$. We propose two options for choosing $\mathbf{x}^*$: 1) minimizing the posterior variance of the integral $\mathbb{V}(Z|\mathcal{D})$ given any new sample,

$$\mathbf{x}^* = \mathrm{argmin}_{\mathbf{x}}\mathbb{V}(Z|\mathcal{D},\mathbf{x}) = \mathrm{argmin}_{\mathbf{x}} Z_B^\top \widetilde{\Sigma}(\boldsymbol{\omega},\mathbf{x})Z_B \quad (27)$$

where $\widetilde{\Sigma}(\boldsymbol{\omega},\mathbf{x})$ is formed by expanding $B_{1:N}$ to $B_{1:N+1}(\mathbf{x})$ in (22); 2) minimizing the uncertainty of the integrand $f(\mathbf{x})p(\mathbf{x})$ instead, by targeting

$$\mathbf{x}^* = \mathrm{argmax}_{\mathbf{x}} \widetilde{\sigma}^2(\mathbf{x},\mathbf{x})p(\mathbf{x})^2 \quad (28)$$

Minimization in (27) involves updating the Fourier matrix $B_{1:N}$ and matrix inversion for each candidate $\mathbf{x}$, while considerable computational savings are realized with (28) though at the cost of the rate of convergence. For (28) we are only calculating the diagonal of the integral covariance and picking the sample which contributes the most. In our experiment, we use the latter for active sampling due to its computational efficiency.

As an illustration, we again examine the ASD model for the Bayesian regression in the previous section and estimate the marginal likelihood by GPBQ with and without Hessian information. Instead of evaluating the log marginal likelihood, we only evaluate the likelihood and place a Gaussian prior over it. The purpose is to marginalize over the hyper-parameters $\boldsymbol{\theta} = \{r, l\}$ and approximate the integral of the likelihood function. Fig. 4 d) indicates the averaged error (absolute value difference) between expected integration calculated by the posterior mean of integral equation and the true integral value. GPBQhess reaches the true value faster than GPBQ. Therefore, derivative information improves both the active sampling and the accuracy of the posterior mean of the integral for Bayesian quadrature.

## 6. Discussion

We have presented novel methods of BO and BQ with gradient- and Hessian-enhanced GPs, and two adjustments in both original and spectral domains are provided to alleviate the ill-conditioning issues for practical use. While derivative-enhanced GPs have been discussed previously (Lockwood and Anitescu, 2012; Dalbey, 2013; Banerjee and Gelfand, 2006; Solak et al., 2003; Wu et al., 2017), there have been no prior attempts at incorporating Hessian information and dealing with the practical ill-conditioning problem caused by the high order derivatives. Prior work has demonstrated both the utility and challenges associated with using observations of function derivatives with GPs (Lockwood and Anitescu, 2012; Dalbey, 2013; Banerjee and Gelfand, 2006; Solak et al., 2003), although previous methods employed different, ad hoc methods to handle the conditioning problem. For example, Osborne et al. (2009) proposed that the conditioning of the covariance is *improved* by the use of derivatives given the fact that, relative to function observations, derivative observations are more weakly correlated with each other. Riihimäki and Vehtari (2010) presented a similar dual-space derivation of the derivative-enhanced GP to our own (Sec. 3), however, they did so assuming that the target function is monotonic. The monotonicity assumption allowed them to handle ill-conditioning with an inequality constraint on the function's derivative, whereas our setting is more general. Dalbey (2013) handled the ill-conditioning by ranking observations based on the how much information they contributed and throwing out the less informative observations. Our methods handle the ill-conditioning problem, even for the use of Hessian observations when the ill-conditioning can be even worse, without use of ad hoc assumptions or removing data. Furthermore, while the use of derivative information has also been developed in the universal Kriging literature (Lockwood and Anitescu, 2012) and Bayesian "wombling" (Banerjee and Gelfand, 2006), the practical use of derivatives for BO or BQ has not been fully realized.

## 7. Conclusion

In conclusion, we have presented an efficient framework for GP-based Bayesian optimization and quadrature in the case where a function's first and second derivatives may be evaluated along with the function itself. The inclusion of higher order derivatives incurs a severe ill-conditioning issue for the kernels prohibiting practical uses. We provided insights into the problem, and proposed a rescaling method for the original kernel and a spectral kernel representation as effective resolutions. We presented applications to BO and BQ using both unimodal and multimodal functions and evidence optimization, and showed that our derivative-enhanced GP displayed faster uncertainty shrinkage, more efficient hyper-parameter sampling, and faster convergence to a global optimum over the standard GPBO and GPBQ.

## References

Sudipto Banerjee and Alan E Gelfand. Bayesian wombling: Curvilinear gradient assessment under spatial process models. *Journal of the American Statistical Association*, 101(476):1487–1501, 2006.

Sudipto Banerjee, Alan E Gelfand, and CF Sirmans. Directional rates of change under spatial process models. *Journal of the American Statistical Association*, 98(464):946–954, 2003.

Salomon Bochner. *Lectures on Fourier Integrals.(AM-42)*, volume 42. Princeton University Press, 2016.

Olivier Chapelle, Vladimir Vapnik, Olivier Bousquet, and Sayan Mukherjee. Choosing multiple parameters for support vector machines. *Machine learning*, 46(1-3):131–159, 2002.

Keith R Dalbey. Efficient and robust gradient enhanced kriging emulators. Technical report, Sandia Technical Report 2013-7022, 2013.

Justin Domke. Generic methods for optimization-based modeling. In *AISTATS*, volume 22, pages 318–326, 2012.

Alexander Forrester, Andras Sobester, and Andy Keane. *Engineering design via surrogate modelling: a practical guide*. John Wiley & Sons, 2008.

Tom Gunter, Michael A Osborne, Roman Garnett, Philipp Hennig, and Stephen J Roberts. Sampling for inference in probabilistic models with fast bayesian quadrature. In *Advances in Neural Information Processing Systems*, pages 2789–2797, 2014.

James Hensman and Neil D Lawrence. Nested variational compression in deep gaussian processes. *arXiv preprint arXiv:1412.1370*, 2014.

Donald R Jones. A taxonomy of global optimization methods based on response surfaces. *Journal of global optimization*, 21(4):345–383, 2001.

Jan Larsen, Claus Svarer, Lars Nonboe Andersen, and Lars Kai Hansen. Adaptive regularization in neural network modeling. In *Neural Networks: Tricks of the Trade*, pages 113–132. Springer, 1998.

Miguel Lázaro-Gredilla, Joaquin Quiñonero-Candela, Carl Edward Rasmussen, and Aníbal R Figueiras-Vidal. Sparse spectrum gaussian process regression. *The Journal of Machine Learning Research*, 11:1865–1881, 2010.

Brian A Lockwood and Mihai Anitescu. Gradient-enhanced universal kriging for uncertainty propagation. *Nuclear Science and Engineering*, 170(2):168–195, 2012.

Dougal Maclaurin, David Duvenaud, and Ryan P Adams. Gradient-based hyperparameter optimization through reversible learning. *arXiv preprint arXiv:1502.03492*, 2015.

Jonas Mockus. Application of bayesian approach to numerical methods of global and stochastic optimization. *Journal of Global Optimization*, 4(4):347–365, 1994.

Jonas Mockus, Vytautas Tiesis, and Antanas Zilinskas. The application of bayesian methods for seeking the extremum. *Towards Global Optimization*, 2(117-129):2, 1978.

Michael Osborne, Roman Garnett, Zoubin Ghahramani, David K Duvenaud, Stephen J Roberts, and Carl E Rasmussen. Active learning of model evidence using bayesian quadrature. In *Advances in Neural Information Processing Systems*, pages 46–54, 2012.

Michael A Osborne, Roman Garnett, and Stephen J Roberts. Gaussian processes for global optimization. In *3rd international conference on learning and intelligent optimization (LION3)*, pages 1–15, 2009.

Christopher J Paciorek. Bayesian smoothing with gaussian processes using fourier basis functions in the spectralgp package. *Journal of statistical software*, 19(2):nihpa22751, 2007a.

Christopher J Paciorek. Computational techniques for spatial logistic regression with large data sets. *Computational statistics & data analysis*, 51(8):3631–3653, 2007b.

Carl Rasmussen and Chris Williams. *Gaussian Processes for Machine Learning*. MIT Press, 2006.

Jaakko Riihimäki and Aki Vehtari. Gaussian processes with monotonicity information. In *International Conference on Artificial Intelligence and Statistics*, pages 645–652, 2010.

J Andrew Royle and Christopher K Wikle. Efficient statistical mapping of avian count data. *Environmental and Ecological Statistics*, 12(2):225–243, 2005.

Jasper Snoek, Hugo Larochelle, and Ryan P Adams. Practical bayesian optimization of machine learning algorithms. In *Advances in Neural Information Processing Systems*, pages 2951–2959, 2012.

E. Solak, R. Murray-smith, W. E. Leithead, D. J. Leith, and Carl E. Rasmussen. Derivative observations in gaussian process models of dynamic systems. In S. Becker, S. Thrun, and K. Obermayer, editors, *Advances in Neural Information Processing Systems 15*, pages 1057–1064. MIT Press, 2003. URL http://papers.nips.cc/paper/2287-derivative-observations-in-gaussian-process-models-of-dynamic-systems.pdf.

Niranjan Srinivas, Andreas Krause, Sham M Kakade, and Matthias Seeger. Gaussian process optimization in the bandit setting: No regret and experimental design. *arXiv preprint arXiv:0912.3995*, 2009.

Ziyu Wang, Masrour Zoghi, Frank Hutter, David Matheson, and Nando De Freitas. Bayesian optimization in high dimensions via random embeddings. In *Proceedings of the Twenty-Third international joint conference on Artificial Intelligence*, pages 1778–1784. AAAI Press, 2013.

Christopher K Wikle. Spatial modeling of count data: A case study in modelling breeding bird survey data on large spatial domains. *Chapman and Hall*, pages 199–209, 2002.

Jian Wu, Matthias Poloczek, Andrew G Wilson, and Peter Frazier. Bayesian optimization with gradients. In *Advances in Neural Information Processing Systems*, pages 5273–5284, 2017.

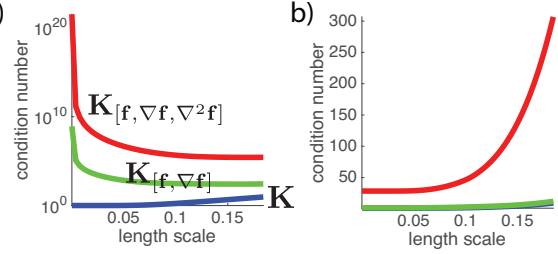

Figure 5: a) Condition number of the Matérn 5/2 covariance matrix $\mathbf{K}$ for a 1D function after 100 observations spaced 0.2 unit apart, as a function of length scale $\delta$. b) Condition number after applying the rescaling trick (note the different scales of the y-axes).

## Appendix

### Analyses and applications for the Matérn 5/2 kernel

The Matérn covariance is commonly used to define the statistical covariance between measurements made at two points that are $d$ units distant from each other. A general mathematical definition is given by

$$k(\mathbf{x}, \mathbf{x}') = \sigma^2 \frac{2^{1-\nu}}{\Gamma(\nu)} \left(\sqrt{2\nu}\frac{d}{\rho}\right)^\nu K_\nu\left(\sqrt{2\nu}\frac{d}{\rho}\right), \quad d = ||\mathbf{x} - \mathbf{x}'|| \qquad (29)$$

where $\Gamma$ is the gamma function, $K_\nu$ is the modified Bessel function of the second kind, and $\rho$ and $\nu$ are non-negative parameters of the covariance. A Gaussian process with Matérn kernel has sample paths that are $\lceil \nu - 1 \rceil$ times differentiable. We are mainly interested in the Matérn 5/2 kernel for $\nu = 5/2$, which is thus twice differentiable for the function sample and 4-times differentiable for the covariance, rendering the appropriate order of derivatives for the kernel when incorporating the gradient and Hessian information.

Since the Matérn kernel also includes the $\exp(||\mathbf{x} - \mathbf{x}'||/\delta)$ term, it is bothered by the ill-conditioning issue with small length scale $\delta$ values as well (Fig. 1 a)). Therefore the rescaling trick is applicable to the Matén 5/2 kernel and other Matén kernels. Fig. 5 shows the alleviated conditions with the rescaling method.

From Fig. 5 a), we note that the Matérn kernel doesn't have a severe ill-conditioning issue when $\delta$ is large, different from the SE kernel. It is because that its spectral form decays in polynomial, where the frequencies for the SE kernel decay to very small values exponentially. Therefore its condition number is still handleable in the smooth regime. But even so, the spectral representation still provides an effective validation tool for manually derived kernels. Differentiating the SE kernel up to the $4^{th}$ or even higher order w.r.t. each dimension is conceptually prohibitive and requires a huge amount of onerous labor intensive work, which easily leads to an inaccurate derivation. However, with the spectral representation, we can effortlessly validate the manually derived kernels with derivatives, calculating only a diagonal matrix and differentiations of constant Fourier bases for stationary kernels. The accuracy-control approximation provides fundamental conditions for a precise validation. Fig. 6 presents comparisons of $\mathbf{K}$, $\mathbf{K}_g = \mathbf{K}_{[\mathbf{f}, \nabla \mathbf{f}]}$ and $\mathbf{K}_H = \mathbf{K}_{[\mathbf{f}, \nabla \mathbf{f}, \nabla^2 \mathbf{f}]}$ for the 1D, 2D and 3D SE kernel via primal and dual implementations. For the Matérn 5/2 kernel, the standard covariance function is not factorizable across dimensions which renders a tough situation for the spectral formulation. Here, we propose a factorizable Matérn 5/2 kernel which is the product of a 1D Matérn kernel along each dimension. We only discard the correlations among dimensions but still encourage the smoothness.

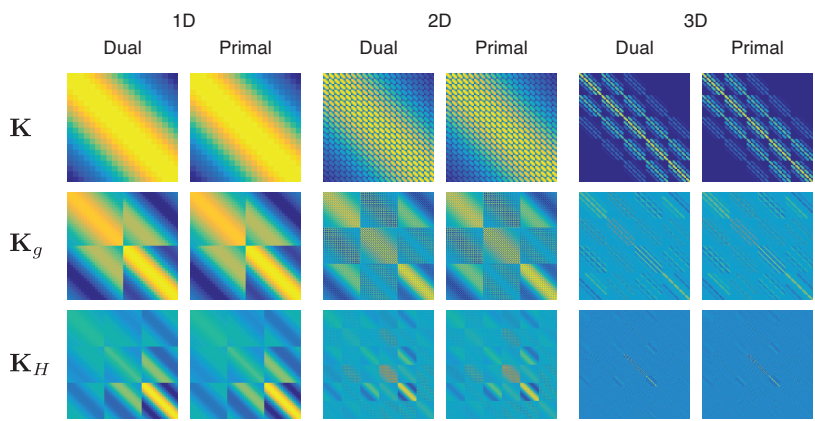

Figure 6: SE kernel

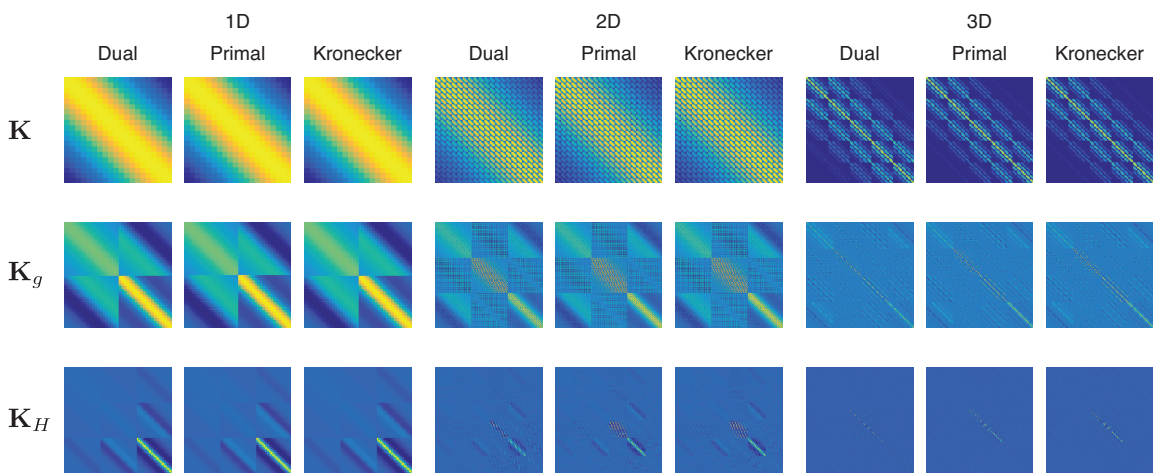

Figure 7: Matérn 5/2 kernel

Fig. 7 presents the 1D, 2D and 3D cases for the Matérn kernel with the standard dual form, the spectral form (primal) and the factorizable dual form (kronecker), respectively. It is apparent that the spectral representations approximate the true kernels very precisely.

To testify its performance on the function evaluations, we also run GPBO methods with the Matérn 5/2 kernel on Rosenbrock, Branin and Shubert functions. Fig. 8 presents the maximal function value and the minimal distance to the true optimal location.

We don't compare with GPBOhess for the Matérn 5/2 kernel due to its unique ill-conditioning issue. As known, the function samples from the Matérn 5/2 kernel are only twice differentiable, which indicates a relatively steeper change around the origin for its second order derivative. This induces big values on the diagonal of the kernel matrices for gradient and Hessian. Therefore, the Matérn 5/2 kernel with Hessian gets more rapid ill-conditioning than its gradient version.

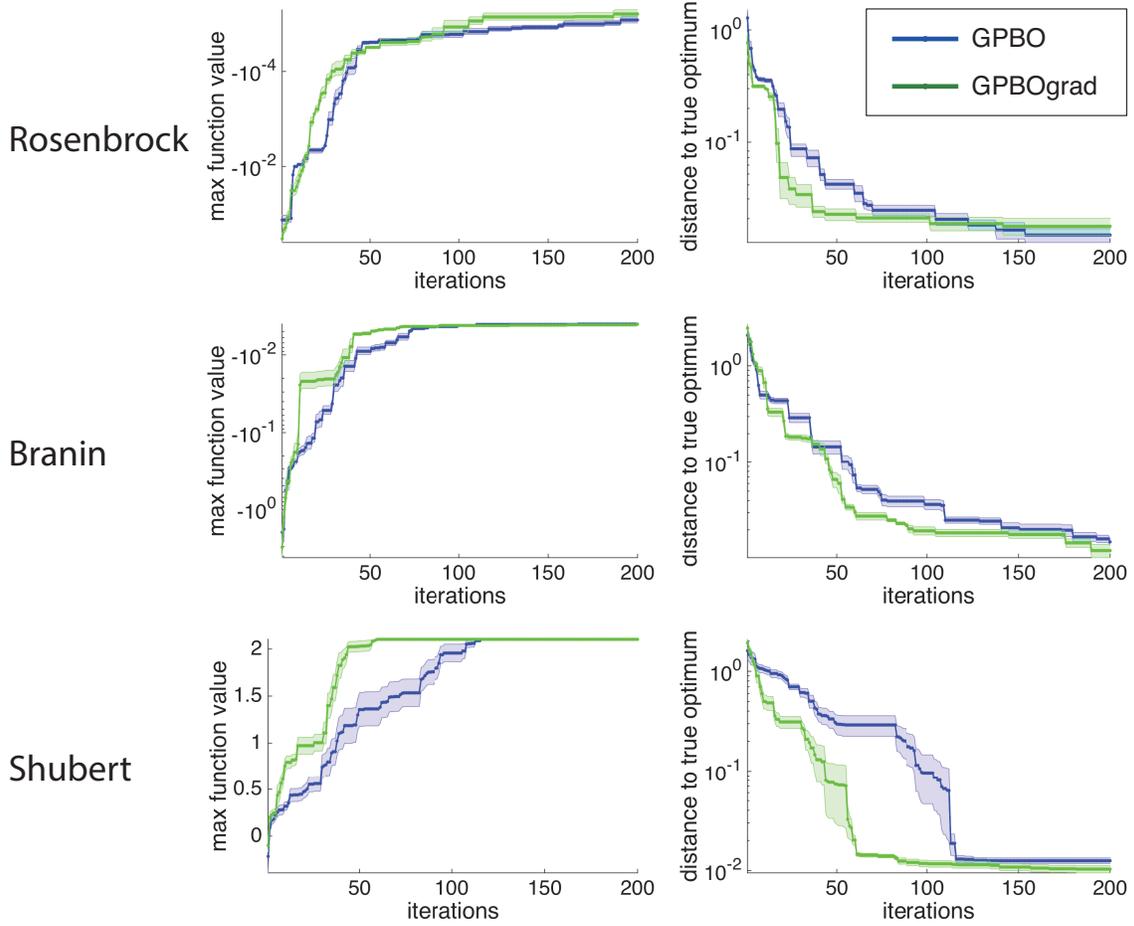

Figure 8: Performance of GP, GPgrad and GPhess for Bayesian optimization with the UCB acquisition function for three functions.